\documentclass[10pt, conference, compsocconf]{IEEEtran}

\ifCLASSINFOpdf
\else
\fi

\usepackage{graphicx}
\usepackage[table,xcdraw]{xcolor}
\usepackage[colorinlistoftodos]{todonotes}
\usepackage{verbatim}
\usepackage{url}
\usepackage{footmisc}
\usepackage{textcomp} 

\begin{document}

\title{Expanding a robot\textquotesingle s life: Low power object recognition via FPGA-based DCNN deployment}

\author{
    \IEEEauthorblockN{
        Panagiotis G. Mousouliotis, Konstantinos L. Panayiotou, Emmanouil G. Tsardoulias, \\Loukas P. Petrou, Andreas L. Symeonidis
    }
    \IEEEauthorblockA{
    	School of Electrical and Computer Engineering, Faculty of Engineering,\\
        Aristotle University of Thessaloniki, Thessaloniki 54124, Greece
    }
}

\maketitle

\begin{abstract}
FPGAs are commonly used to accelerate domain-specific algorithmic implementations, as they can achieve impressive performance boosts, are reprogrammable and exhibit minimal power consumption. In this work, the SqueezeNet DCNN is accelerated using an SoC FPGA in order for the offered object recognition resource to be employed in a robotic application. Experiments are conducted to investigate the performance and power consumption of the implementation in comparison to deployment on other widely-used computational systems.
\end{abstract}


\IEEEpeerreviewmaketitle


\section{Introduction}
\label{sec:intro}

Object recognition is considered one of the most fundamental abilities an autonomous robot should exhibit. Traditionally, object recognition is performed via feature employment (e.g. SIFT, SURF, ORB), nevertheless current advances in deep learning have crowned DCNNs (Deep Convolutional Neural Networks) as the kings of the field, since they have outperformed any previous approaches in terms of recognition accuracy. Leaving aside their high training times, one of the main drawbacks of DCNNs is the amount of convolution operations needed to perform a forward pass in the NN, i.e. to execute a classification operation given an image as input. It is true that modern CPUs and/or GPUs are quite capable of executing such operations in high frequencies, but unfortunately power consumption is rather high, which has a direct impact on the battery of a robot (specially in the case of drones).

The current paper presents a real-world robotics application and an evaluation of the work in \cite{SJ}. Specifically, a distributed robotic architecture is proposed, which embeds a DCNN deployed in a System on Chip (SoC) FPGA, namely the SqueezeNet v1.1 DCNN (SqN). SqN is accelerated by the SqueezeJet FPGA accelerator (SqJ) \cite{SJ} and is remotely executed on a Xilinx Zynq Platform. This approach achieves comparable performance to common CPUs but with less power consumption, while it also provides remotely accessible resources for object recognition tasks.


\section{State of the art}
\label{sec:sota}

Although there is not much research work on presenting full FPGA-based DCNN systems for robotic applications, many approaches exist where FPGAs are used either in robotics or for vision applications (including DCNNs). As far as FPGA employment in robots is concerned, work in \cite{SMG/FPGA} has implemented the scan-matching genetic-based SMG-SLAM algorithm on Xilinx Virtex-5, achieving almost 15 times faster iteration times in comparison to the original algorithm. Also, interesting control and navigation-oriented implementations have been proposed, such as \cite{CloControl}, which investigates an FPGA-based PID motion control system for small, self-adaptive systems and \cite{ICControl}, which removes a servo control loop from the digital signal processor (DSP) and implements a high-speed servo loop in an FPGA.

Besides the robotics domain, several FPGA implementations of vision-related algorithms exist (including DCNNs), which are described as ``robotics-suitable'', however have not been tested in real-life conditions. For example, in \cite{VFilter}, a parallel implementation of low-level image filtering was created in an FPGA-based system, whereas in \cite{SURF/FPGA} and \cite{FPGA/ORB} common feature extractors like SURF, Harris-Stephens corner detector or ORB were implemented in FPGAs.

Regarding NN implementations, several DCNN FPGA acceleration approaches exist, such as in \cite{FPGA/CNN} where a scalable hardware architecture to implement large-scale CNNs and state-of-the-art multi-layered artificial vision systems is presented, \cite{FPGA/CNN2} where an efficient implementation for accelerating DCNNs on a mobile platform (Xilinx Zynq-7000) in a pipelined manner, \cite{CONV/FP}, which presents an FPGA implementation of CNN designed for addressing portability and power efficiency and \cite{DEEPER}, where a CNN accelerator design on embedded FPGA for Image-Net large-scale image classification is proposed. Finally, architectures such as Angel-Eye \cite{ANGEL} offer quantization strategies and compilation tools taking into account requirements on memory, computation, and the flexibility of the system.

In our work, we propose offloading the execution of the SqN DCNN on the SoC FPGA of the Xilinx ZC702 board using the SqJ accelerator, invoked from an application deployed in the ARM Cortex-A9 core of the same SoC FPGA. The inference of the SqN DCNN is exposed via a single Microservice (uService) to a robotics controller in order to provide remote access, acting as a stand-alone physical node capable of performing object recognition tasks.


\section{Implementation}
\label{sec:implementation}

\subsection{System architecture}
\label{subsec:general}

\begin{figure}[h]
  \centering
  \includegraphics[width=0.5\textwidth]{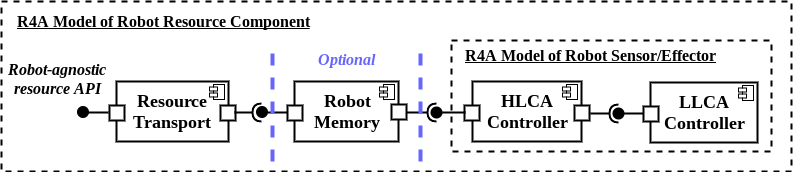}
  \caption{Simplified R4A architecture conceptualization}
  \label{r4a}
\end{figure}

\begin{figure*}[th!]
  \centering
  \includegraphics[width=1\textwidth]{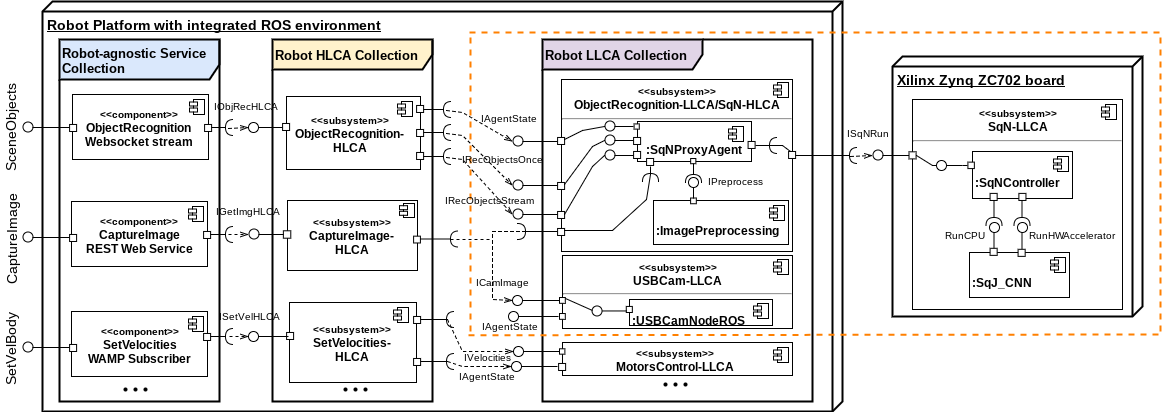}
  \caption{Overall system architecture}
  \label{arch}
\end{figure*}

System design and deployment processes follow the R4A methodology and system architecture \cite{R4A} which exposes robotic (or device) resources in a unified and agnostic way. In figure \ref{r4a}, each device contains an LLCA (\emph{Low Level Core Agent}) that includes hardware/device drivers and exposes raw functionalities, and an HLCA (\emph{High Level Core Agent}) that can manage the operation mode of the respective LLCA, perform pre/post-processing, abstract functionalities by making them vendor-agnostic and expose them via over-network transports. As far as deployment is concerned, LLCA lives within the physical boundaries of each robot/device, while relaxed robot/device boundaries are applied for the other components within the R4A architecture (HLCA, Robot Memory and Resource Transports). The latter can be deployed either in-robot or on a remote node or platform.

Conforming to the R4A architecture, an LLCA has been developed and deployed on the Xilinx Zynq ZC702 embedded board, exposing the inference operation of SqN via a single uService, in order to provide remote access to object recognition tasks. The uService is implemented via a headless protocol for transmitting raw image data over a plain TCP connection, whereas response data consists of the top-5 object classes (ILSVRC 2012 dataset\footnote{\url{http://www.image-net.org/challenges/LSVRC/2012/}}) along with their probability of certainty. As denoted in figure \ref{arch}, the robotic resources layer is located on the physical node executed in the robotic platform, and includes several LLCA instances which offer functionalities to the application layer.

SqN-HLCA is deployed on the robotic platform, acting as a proxy to the SqN-LLCA. It consists of a controller (SqNProxyAgent) for managing the internal state and a stack of preprocessing operations that are applied to the images published by a ROS Node \cite{ROS}, implementing the HW interface of usb cameras (UsbCamNodeROS). SqN-HLCA provides three interfaces: a ROS-Service for setting the internal state of the SqN-HLCA, a ROS-Publisher for streaming the classification results, and a ROS-Service for one-shot object recognition operations. An interesting observation is that SqN-HLCA is at the same time an LLCA of the robot since it offers raw resources to the robot controller (ObjectRecognition-LLCA).
In figure \ref{arch}, the orange dashed lines indicate the boundary limits of the current implementation. To further expose robot functionalities in an agnostic manner, for each LLCA a corresponding HLCA must ultimately be deployed.

\subsection{FPGA accelerator implementation}
\label{subsec:fpga}

The tools employed for the FPGA accelerator development are: (1) the \emph{Xilinx Vivado HLS} (VHLS), which allows the seamless conversion of high-level programming language source code (such as C/C++) to efficient hardware description language (HDL) code that can then be synthesized and mapped to FPGA devices and (2) the \emph{Xilinx SDSoC Environment}, which raises the abstraction level of FPGA application development even more compared to VHLS by providing C/C++/OpenCL HLS capabilities, easy directive driven interface synthesis and automated (Standalone, Linux) OS image generation for the developed application.

SqJ accelerates all but the first SqN convolutional layer due to its architectural differences compared to all the other convolutional SqN layers, whilst a dedicated accelerator is designed to model the first layer. The common characteristics of the SqJ-accelerated layers are a stride equal to 1, an input channel dimension with a greatest common divisor (GCD) equal to 16, and an output channel dimension which is divisible by a power of 2. Since the HLS compiler cannot unroll loops with variable-length bounds, the input channel GCD is used as a design parameter of a pipelined multiply-accumulate unit (MAC-16) which performs 16 MACs in every clock cycle of its operation. Using the output channel characteristic, the MAC-16 unit is replicated $2^n~ (n=2, 3, ...)$ times and the resulting architecture is used to concurrently calculate $2^n$ output channels. Since the parallelism exploitation is focused on the input and output channel dimensions, SqJ can support both $1\times1$ and $3\times3$ kernel sizes.

\begin{figure}[h!]
  \centering
  \includegraphics[width=0.5\textwidth]{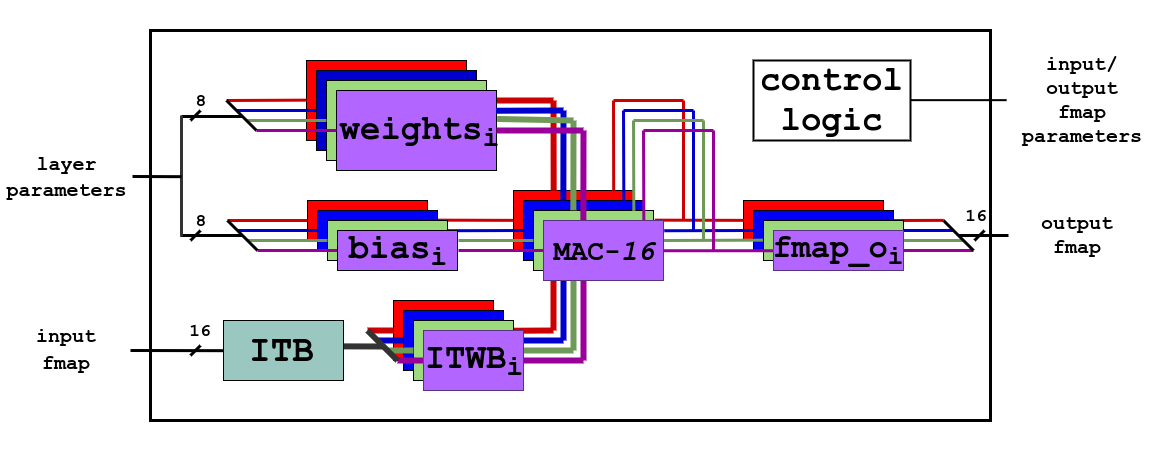}
  \caption{SqJ Block Diagram}
  \label{sqj}
\end{figure}

The Ristretto tool \cite{RISTR} is used to squeeze the SqN parameters (weights, bias) to 8 bits and the SqN future maps (fmaps) to 16 bits with 0.88\% top-5 accuracy loss without performing any fine-tuning. Parameter and fmaps quantization aims at: (1) making the SqJ design smaller, requiring much fewer FPGA resources than the floating-point design and fitting into low-end FPGA devices, (2) storing the SqN parameters in Block RAM (BRAM) FPGA resources since they are used for the calculation of every output multi-channel pixel of the output fmap, avoiding unnecessary memory accesses which introduce additional latency and power consumption. Additionally to the parameter buffering design choice, an input fmap tile buffer (ITB) and an input fmap tile buffer window (ITBW) are used. After the initialization of these two buffers, SqJ consumes and produces data pixel-by-pixel (in SqJ jargon, a pixel consists of all the channels at a specific $(x, y)$ location in the fmap volume \cite{SJ}). The ITB and ITBW are designed to take advantage of the spatial locality of the convolution input data and minimize unnecessary data movement. Figure \ref{sqj} shows, for simplicity, SqJ implemented with 4 MAC-16 units. In this work, an SqJ with 8 MAC-16 units is used and it runs at 100MHz. The FPGA resource utilization of SqJ is shown in Table \ref{table1}.

\section{Experiments / Results}
\label{sec:experiments}

Two boards, an Intel NUC and an Ultrabook, both with Intel low power processors, namely the Intel® Core™ i3-7100U@2.4GHz and the Intel® Core™ i5-3337U@1.8GHz, are used as alternative computation nodes to the Xilinx XC702. Specifically, the SqN application is offloaded on 4 different computation node configurations acquiring latency and power measurements, as shown in Table \ref{table2}. Additionally, Table \ref{table2} reports local SqN runs and provides latency results per CNN layer. All the latency results are the average value of 100 iterations and the power consumption results are acquired: (1) using Intel PCM\footnote{\label{note1}\url{https://www.intel.com/software/pcm}} while the computation node serves 1000 image recognition requests, in the case of the Intel CPUs, and (2) using Xilinx XPE\footnote{\label{note2}\url{https://www.xilinx.com/products/technology/power/xpe.html}} in the case of Xilinx XC702. SqN consists of single-threaded 32-bit floating point precision C++ function accelerated with single-instruction multiple-data (SIMD) instruction set extensions (Intel AVX, ARM NEON) and executed on a single core of the target computation nodes. 

Results report a remote application performance in frames per second (fps) of (see \textbf{End-To-End} observation in Table \ref{table2}) 4.16fps for the i3 core, 2.92fps for the i5 core, 0.19fps for the ARM core, and 2.62fps for the ARM+SqJ cores (SqJ executes only the Conv+Fire layers). The \textbf{Total Conv+Fire} and \textbf{Chip Power} results indicate that the ARM+SqJ configuration is slightly faster on convolution operation execution than the i5 core and 2.68 times more power efficient. Although the Intel i3 core operates in higher frequency than the i5, it consumes less power due to the newer technology used (14nm vs 22nm) and architecture improvements. Finally, SqJ provides to the remote SqN application a 13.487 times speedup compared to the ARM-only configuration.

\begin{table}[]
\centering
\caption{SqJ Resource Utilization on the XC7Z020 SoC FPGA}
\label{table1}

\scalebox{0.8}{
\begin{tabular}{|
>{\columncolor[HTML]{EFEFEF}}l |r|r|r|}
\hline
\textbf{Resource}        & \cellcolor[HTML]{EFEFEF}\textbf{Utilization} & \cellcolor[HTML]{EFEFEF}\textbf{Available} & \cellcolor[HTML]{EFEFEF}\textbf{Utilization \%} \\ \hline
\textbf{LUT}    & 20148                                        & 53200                                      & 37.87                                        \\ \hline
\textbf{LUTRAM} & 1273                                         & 17400                                      & 7.32                                       \\ \hline
\textbf{FF}     & 29568                                        & 106400                                     & 27.55                                        \\ \hline
\textbf{BRAM}   & 134.5                                        & 140                                        & 96.07                                        \\ \hline
\textbf{DSP}    & 192                                          & 220                                        & 87.27                                        \\ \hline
\end{tabular}
}
\end{table}  

\begin{table}[]
\centering
\caption{SqN Remote/Local Application Results}
\label{table2}

\scalebox{0.70}{
\begin{tabular}{lrrrr}
\hline
\multicolumn{5}{|c|}{\cellcolor[HTML]{AAAAAA}\textbf{Experimental platforms}}                                                                                                                                                                                                                                                                                                                                                                                                                                                                                                                                                       \\
\hline
\multicolumn{1}{|l|}{}                                                                                                   & \multicolumn{1}{c|}{\cellcolor[HTML]{EFEFEF}\textbf{\begin{tabular}[c]{@{}c@{}}NUC\\ Intel-i3@2.4GHz\end{tabular}}} & \multicolumn{1}{c|}{\cellcolor[HTML]{EFEFEF}\textbf{\begin{tabular}[c]{@{}c@{}}Ultrabook\\ Intel-i5@1.8GHz\end{tabular}}} & \multicolumn{1}{c|}{\cellcolor[HTML]{EFEFEF}\textbf{\begin{tabular}[c]{@{}c@{}}ZC702\\ ARM@667MHz\end{tabular}}} & \multicolumn{1}{c|}{\cellcolor[HTML]{EFEFEF}\textbf{\begin{tabular}[c]{@{}c@{}}ZC702\\ ARM@667MHz\\ SqJ@100MHz\end{tabular}}} \\ \hline

\multicolumn{5}{|c|}{\cellcolor[HTML]{AAAAAA}\textbf{SqN Remote Application Latency Results (ms)}}                                                                                                                                                                                                                                                                                                                                                                                                                                                                                                                                                       \\ \hline

\multicolumn{1}{|l|}{\cellcolor[HTML]{EFEFEF}\textbf{\begin{tabular}[c]{@{}l@{}}Img Preprocessing\end{tabular}}}     & \multicolumn{1}{r|}{10.0961}                                                                                        & \multicolumn{1}{r|}{10.4999}                                                                                              & \multicolumn{1}{r|}{10.0174}                                                                                     & \multicolumn{1}{r|}{10.3446}                                                                                                   \\ \hline
\multicolumn{1}{|l|}{\cellcolor[HTML]{EFEFEF}\textbf{SqN Inference}}                                                     & \multicolumn{1}{r|}{181.8990}                                                                                       & \multicolumn{1}{r|}{285.5170}                                                                                             & \multicolumn{1}{r|}{5057.6200}                                                                                   & \multicolumn{1}{r|}{323.3100}                                                                                                  \\ \hline
\multicolumn{1}{|l|}{\cellcolor[HTML]{EFEFEF}\textbf{\begin{tabular}[c]{@{}l@{}}Net Transfer\end{tabular}}}        & \multicolumn{1}{r|}{58.2517}                                                                                        & \multicolumn{1}{r|}{57.3641}                                                                                              & \multicolumn{1}{r|}{91.6487}                                                                                     & \multicolumn{1}{r|}{58.4605}                                                                                                   \\ \hline
\multicolumn{1}{|l|}{\cellcolor[HTML]{EFEFEF}\textbf{End-To-End}}                                                        & \multicolumn{1}{r|}{\textbf{240.1507}}                                                                                       & \multicolumn{1}{r|}{\textbf{342.8810}}                                                                                             & \multicolumn{1}{r|}{\textbf{5149.2687}}                                                                                   & \multicolumn{1}{r|}{\textbf{381.7705}}                                                                                                  \\ \hline
\multicolumn{1}{|l|}{\cellcolor[HTML]{EFEFEF}\textbf{Total}}                                                             & \multicolumn{1}{r|}{250.2468}                                                                                       & \multicolumn{1}{r|}{353.3810}                                                                                             & \multicolumn{1}{r|}{5159.2861}                                                                                   & \multicolumn{1}{r|}{392.1151}                                                                                                  \\ \hline

\multicolumn{5}{|c|}{\cellcolor[HTML]{AAAAAA}\textbf{SqN Local Application Per Layer Latency Results (ms)}}                                                                                                                                                                                                                                                                                                                                                                                                                                                                                                                                              \\ \hline

\multicolumn{1}{|l|}{\cellcolor[HTML]{EFEFEF}\textbf{1:Conv}}                                                            & \multicolumn{1}{r|}{25.5531}                                                                                        & \multicolumn{1}{r|}{35.6304}                                                                                              & \multicolumn{1}{r|}{297.3461}                                                                                    & \multicolumn{1}{r|}{26.4994}                                                                                                   \\ \hline
\multicolumn{1}{|l|}{\cellcolor[HTML]{EFEFEF}\textbf{2:Maxpool}}                                                         & \multicolumn{1}{r|}{2.2457}                                                                                         & \multicolumn{1}{r|}{3.4679}                                                                                               & \multicolumn{1}{r|}{28.7091}                                                                                     & \multicolumn{1}{r|}{22.7482}                                                                                                   \\ \hline
\multicolumn{1}{|l|}{\cellcolor[HTML]{EFEFEF}\textbf{3:Fire}}                                                            & \multicolumn{1}{r|}{16.6766}                                                                                        & \multicolumn{1}{r|}{25.4867}                                                                                              & \multicolumn{1}{r|}{446.0529}                                                                                    & \multicolumn{1}{r|}{32.7412}                                                                                                   \\ \hline
\multicolumn{1}{|l|}{\cellcolor[HTML]{EFEFEF}\textbf{4:Fire}}                                                            & \multicolumn{1}{r|}{17.8092}                                                                                        & \multicolumn{1}{r|}{26.8687}                                                                                              & \multicolumn{1}{r|}{474.0225}                                                                                    & \multicolumn{1}{r|}{34.8575}                                                                                                   \\ \hline
\multicolumn{1}{|l|}{\cellcolor[HTML]{EFEFEF}\textbf{5:Maxpool}}                                                         & \multicolumn{1}{r|}{1.5101}                                                                                         & \multicolumn{1}{r|}{2.1909}                                                                                               & \multicolumn{1}{r|}{27.3655}                                                                                     & \multicolumn{1}{r|}{18.0697}                                                                                                   \\ \hline
\multicolumn{1}{|l|}{\cellcolor[HTML]{EFEFEF}\textbf{6:Fire}}                                                            & \multicolumn{1}{r|}{14.167}                                                                                         & \multicolumn{1}{r|}{20.6089}                                                                                              & \multicolumn{1}{r|}{450.0639}                                                                                    & \multicolumn{1}{r|}{17.8422}                                                                                                   \\ \hline
\multicolumn{1}{|l|}{\cellcolor[HTML]{EFEFEF}\textbf{7:Fire}}                                                            & \multicolumn{1}{r|}{15.1649}                                                                                        & \multicolumn{1}{r|}{22.0343}                                                                                              & \multicolumn{1}{r|}{482.4270}                                                                                    & \multicolumn{1}{r|}{19.0028}                                                                                                   \\ \hline
\multicolumn{1}{|l|}{\cellcolor[HTML]{EFEFEF}\textbf{8:Maxpool}}                                                         & \multicolumn{1}{r|}{0.06697}                                                                                        & \multicolumn{1}{r|}{1.0116}                                                                                               & \multicolumn{1}{r|}{14.4056}                                                                                     & \multicolumn{1}{r|}{9.4262}                                                                                                    \\ \hline
\multicolumn{1}{|l|}{\cellcolor[HTML]{EFEFEF}\textbf{9:Fire}}                                                            & \multicolumn{1}{r|}{7.7804}                                                                                         & \multicolumn{1}{r|}{11.1605}                                                                                              & \multicolumn{1}{r|}{258.0127}                                                                                    & \multicolumn{1}{r|}{8.6744}                                                                                                    \\ \hline
\multicolumn{1}{|l|}{\cellcolor[HTML]{EFEFEF}\textbf{10:Fire}}                                                           & \multicolumn{1}{r|}{8.2085}                                                                                         & \multicolumn{1}{r|}{11.6817}                                                                                              & \multicolumn{1}{r|}{273.4767}                                                                                    & \multicolumn{1}{r|}{8.8977}                                                                                                    \\ \hline
\multicolumn{1}{|l|}{\cellcolor[HTML]{EFEFEF}\textbf{11:Fire}}                                                           & \multicolumn{1}{r|}{13.7099}                                                                                        & \multicolumn{1}{r|}{19.3248}                                                                                              & \multicolumn{1}{r|}{497.9448}                                                                                    & \multicolumn{1}{r|}{12.2668}                                                                                                   \\ \hline
\multicolumn{1}{|l|}{\cellcolor[HTML]{EFEFEF}\textbf{12:Fire}}                                                           & \multicolumn{1}{r|}{14.2955}                                                                                        & \multicolumn{1}{r|}{20.0220}                                                                                              & \multicolumn{1}{r|}{517.3455}                                                                                    & \multicolumn{1}{r|}{12.8121}                                                                                                   \\ \hline
\multicolumn{1}{|l|}{\cellcolor[HTML]{EFEFEF}\textbf{13:Conv}}                                                           & \multicolumn{1}{r|}{36.3992}                                                                                        & \multicolumn{1}{r|}{49.9700}                                                                                              & \multicolumn{1}{r|}{1258.8026}                                                                                   & \multicolumn{1}{r|}{49.5907}                                                                                                   \\ \hline
\multicolumn{1}{|l|}{\cellcolor[HTML]{EFEFEF}\textbf{14:Avgpool}}                                                        & \multicolumn{1}{r|}{1.6158}                                                                                         & \multicolumn{1}{r|}{1.5544}                                                                                               & \multicolumn{1}{r|}{5.7776}                                                                                      & \multicolumn{1}{r|}{5.7192}                                                                                                    \\ \hline
\multicolumn{1}{|l|}{\cellcolor[HTML]{EFEFEF}\textbf{15:Softmax}}                                                        & \multicolumn{1}{r|}{0.0277}                                                                                         & \multicolumn{1}{r|}{0.0420}                                                                                               & \multicolumn{1}{r|}{0.2242}                                                                                      & \multicolumn{1}{r|}{0.2255}                                                                                                    \\ \hline
\rowcolor[HTML]{FFFFFF} 
\multicolumn{1}{|l|}{\cellcolor[HTML]{C0C0C0}\textbf{Total Conv+Fire}}                                                   & \multicolumn{1}{r|}{\cellcolor[HTML]{FFFFFF}\textbf{169.7643}}                                                      & \multicolumn{1}{r|}{\cellcolor[HTML]{FFFFFF}\textbf{242.7880}}                                                            & \multicolumn{1}{r|}{\cellcolor[HTML]{FFFFFF}\textbf{4955.4947}}                                                  & \multicolumn{1}{r|}{\cellcolor[HTML]{FFFFFF}\textbf{223.1848}}                                                                 \\ \hline
\rowcolor[HTML]{FFFFFF} 
\multicolumn{1}{|l|}{\cellcolor[HTML]{C0C0C0}\textbf{Total}}                                                             & \multicolumn{1}{r|}{\cellcolor[HTML]{FFFFFF}\textbf{175.23057}}                                                     & \multicolumn{1}{r|}{\cellcolor[HTML]{FFFFFF}\textbf{251.0548}}                                                            & \multicolumn{1}{r|}{\cellcolor[HTML]{FFFFFF}\textbf{5031.9767}}                                                  & \multicolumn{1}{r|}{\cellcolor[HTML]{FFFFFF}\textbf{279.3736}}                                                                 \\\hline
\multicolumn{5}{|c|}{\cellcolor[HTML]{AAAAAA}\textbf{SqN Remote Application CPU/SoC Power Results (Watts)}}                                                                                                                                                                                                                                                                                                                                                                                                                                                                                                                                                      \\ \hline

\multicolumn{1}{|l|}{\cellcolor[HTML]{EFEFEF}\textbf{Technology}}       & \multicolumn{1}{r|}{14nm}                                                                                           & \multicolumn{1}{r|}{22nm}                                                                                                 & \multicolumn{1}{r|}{28nm}                                                                                        & \multicolumn{1}{r|}{28nm}                                          \\ \hline

\multicolumn{1}{|l|}{\cellcolor[HTML]{EFEFEF}\textbf{Chip Power}}       & \multicolumn{1}{r|}{4.1187}                                                                                           & \multicolumn{1}{r|}{5.9883}                                                                                                 & \multicolumn{1}{r|}{1.629}                                                                                        & \multicolumn{1}{r|}{\textbf{2.227}}                                          \\ \hline

\end{tabular}
}
\end{table}


\section{Conclusion / Future Work}
\label{sec:conclusion}
In the context of the current paper, a distributed implementation of the SqueezeNet CNN is proposed, exploiting low power consumption capabilities of FPGA-based embedded systems. The application execution is distributed between the Xilinx XC702 device, and a ROS-enabled node. Performance is expressed in terms of both execution time and power consumption, and results indicate comparable execution times at lower power consumption rates, over common CPUs.
Though, most robotic applications not only require to observe the existence of several objects, but also to localize them in the scene, or even on a global map. For this purpose, SqueezeDet network \cite{SqueezeDet} can be used to support both recognition and multi-object localization tasks.
Furthermore, the SqJ convolutional hardware accelerator could be redesigned to support: (1) Maxpool layers, since they require considerable amount (almost 20\%) of the total inference time on a mobile ARM core, and (2) streaming execution, to avoid memory accesses for fmaps (requires additional BRAM resources).



\end{document}